\documentclass{article}
\usepackage[T1]{fontenc}
\usepackage{array}
\usepackage{rotating}
\usepackage{textcomp}
\usepackage{multirow}
\usepackage{amsmath}
\usepackage{amsthm}
\usepackage{graphicx}
\usepackage[numbers]{natbib}

\makeatletter

\let\SF@@footnote\footnote
\def\footnote{\ifx\protect\@typeset@protect
    \expandafter\SF@@footnote
  \else
    \expandafter\SF@gobble@opt
  \fi
}
\expandafter\def\csname SF@gobble@opt \endcsname{\@ifnextchar[
  \SF@gobble@twobracket
  \@gobble
}
\edef\SF@gobble@opt{\noexpand\protect
  \expandafter\noexpand\csname SF@gobble@opt \endcsname}
\def\SF@gobble@twobracket[#1]#2{}
\providecommand{\tabularnewline}{\\}

  \theoremstyle{definition}
  \newtheorem{defn}{\protect\definitionname}

\usepackage[preprint]{neurips_2019}
\renewcommand{\@notice}{}

\usepackage{url}
\usepackage{amsfonts}
\usepackage{nicefrac}
\usepackage{microtype}

\makeatother

  \providecommand{\definitionname}{Definition}

\begin{document}

\title{Learning Set-equivariant Functions with SWARM Mappings}

\author{Roland Vollgraf\\
Zalando Research\\
Zalando SE\\
10243 Berlin, Germany \\
\texttt{roland.vollgraf@zalando.de}}
\maketitle
\begin{abstract}
In this work we propose a new neural network architecture that efficiently
implements and learns general purpose set-equivariant functions. Such
a function $f$ maps a set of entities $x=\left\{ x_{1},\ldots,x_{n}\right\} $
from one domain to a set of same cardinality $y=f\left(x\right)=\left\{ y_{1},\ldots,y_{n}\right\} $
in another domain regardless of the ordering of the entities. The
architecture is based on a gated recurrent network which is iteratively
applied to all entities individually and at the same time syncs with
the progression of the whole population. In reminiscence to this pattern,
which can be frequently observed in nature, we call our approach \emph{SWARM
}mapping.

Set-equivariant and generally permutation invariant functions are
important building blocks for many state of the art machine learning
approaches. Even in applications where the permutation invariance
is not of primary interest, as to be seen in the recent success of
attention based transformer models \citep{DBLP:conf/nips/VaswaniSPUJGKP17}.
Accordingly, we demonstrate the power and usefulness of SWARM mappings
in different applications. We compare the performance of our approach
with another recently proposed set-equivariant function, the \emph{Set
Transformer} \citep{DBLP:journals/corr/abs-1810-00825} and we demonstrate
that models solely based on SWARM layers gives state of the art results.
\end{abstract}

\section{Introduction}

Permutation invariant transformations have recently attracted growing
attention in the research community. Today, there are numerous deep
learning tasks where data comes in an unordered or non-meaningful
order. Think of, for example, an image based classification task,
where the decision has to be made based on a collection of images.
The order in the data batch often is arbitrary and non-informative,
though the classifier may be sensitive to it. When empirically marginalizing
over the ordering, the sensitivity reflects in the variance of the
classifier. We will demonstrate this effect in a little example below.
But interesting applications are not limited to information pooling
from collections. In principle everywhere where information on a population
of entities is processed - be it to take a decision upon the whole
population or a decision on the individuals that is influenced by
the population - permutation invariant or equivariant functions emerge.
The population can be, as already mentioned, a collection of objects
to classify jointly, the data points in Bayesian experiment, examples
in a few-shot learning setup, and many more. The theory of invariant
functions is well understood. \citep{NIPS2017_6931} have introduced
the notion of \emph{Deep Sets }as learnable set functions. \citep{DBLP:journals/corr/abs-1804-10306}
and \citep{DBLP:journals/corr/abs-1903-01939} study generalizations
of the universal approximation theorem for neural networks for invariant
or equivariant mappings. However, it is not clear if the theoretical
results always provide useful foundation for designing practically
applicable set functions \citep{DBLP:journals/corr/abs-1901-09006}.

In this work, we propose a new approach to set-equivariant functions
that practically works well and efficiently also under circumstances
where approaches inspired by universal approximation theorem do not.
First we will introduce our model, which we call SWARM mappings. We
will then introduce an amortized clustering task as a challenging
performance benchmark. We compare SWARM mappings with other approaches
to set-equivariant functions. Further we demonstrate that SWARM mappings
can also be used in a not equivariant setting by allowing a setup
of a 1-layer transformer architecture for the generation of images

\section{Set-equivariant functions}

We study problems in which an unordered set or \emph{population} of
\emph{entities} is processed simultaneously by a deep neural network.
We use bold face symbols or a notation in parentheses to indicate
the whole population of entities as a matrix $\mathbf{x}$ or $\left\{ x_{1},\ldots,x_{N}\right\} =\left\{ x_{i}\right\} _{i=1\ldots N}\in\mathbb{R}^{d\times N}$
as a set. Whenever there is no ambiguity, we may omit the subscripts
$\left\{ \cdot\right\} {}_{i=1\ldots N}$ for simplicity. Although
the population of entities is a set of vectors, it makes sense to
consider them in arbitrary but fixed order as a matrix. For a set-equivariant
mapping we have to ensure that in can be carried out on an arbitrary
number of entities and their ordering doesn't matter.
\begin{defn}
A function $f\ :\ \mathbb{R}^{d_{x}\times N}\rightarrow\mathbb{R}^{d_{y}\times N}$
is set-equivariant if it is defined for all $N\in\mathbb{N}^{+}$
and for all $\mathbf{x}\in\mathbb{R}^{d_{x}\times N}$ the following
holds

\begin{align}
\pi\left(f\left(\mathbf{x}\right)\right) & =f\left(\pi\left(\mathbf{x}\right)\right)\label{eq:set-equivariant}
\end{align}
for arbitrary permutations of the columns of $\mathbf{x}$, $\pi\left(\mathbf{x}\right):=\left(x_{\pi(i)}\right)_{i=1\ldots N_{E}}$
.
\end{defn}
From equation (\ref{eq:set-equivariant}) it follows directly that
for a repeated application of functions $f=f_{1}\circ\ldots\circ f_{n}$
to be a set-equivariant mapping it is sufficient that every $f_{i}$
is set-equivariant. Thus, we can model arbitrarily complex functions
in a hierarchical structure, just like in any other feed forward neural
network architecture, as long as we provide that all components fulfill
(\ref{eq:set-equivariant}). Apparently, any function that maps entities
individually is trivially set-equivariant. Standard non-linearities
or entity-wise linear or non-linear operations (sometimes referred
to as $1\times1$-convolutions) fall into that category.

However, the family of functions that fulfill the definition is much
richer than this. The simplest non-trivial one is the linear mapping
\begin{align}
\mathbf{y}_{ji} & =\sum_{l=1}^{d_{x}}\sum_{k=1}^{N}\mathbf{W}_{ji,lk}\mathbf{x}_{lk}+\mathbf{b}_{ji}\quad\textrm{{where}} & \mathbf{W}_{ji,lk} & =\begin{cases}
\mathbf{W}_{jl}^{=}, & i=k\\
\mathbf{W}_{jl}^{\neq}, & i\neq k
\end{cases}\ , & \mathbf{b}_{ji} & =\mathbf{b}_{j}\ .\label{eq:set_linear}
\end{align}
In fact, this is equivalent to two linear functions, one operating
on all entities individually ($\mathbf{W}^{=}$) and one working on
all entities summed up ($\mathbf{W}^{\neq}$), the output of which
as well as the bias are shared by all entities. In a feed forward
architecture with several such layers combined with appropriate non-linearities,
significantly non-trivial set functions can be learned. It has been
proven theoretically that such set pooling functions can approximate
arbitrary complex set-equivariant (and -invariant) functions \citep{NIPS2017_6931}.
However, in practical applications, this structure can turn out to
be too limiting \citep{DBLP:journals/corr/abs-1901-09006}. We will
call such layers 'set-linear' layers in the following and include
them in our experiments below.

\section{SWARM Mappings}

\begin{figure}
a)\includegraphics[bb=0bp 0bp 329bp 235bp,width=0.3\textwidth]{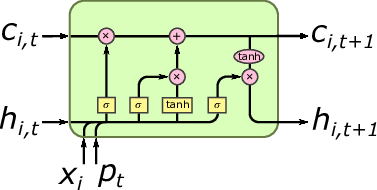}b)\includegraphics[width=0.7\textwidth]{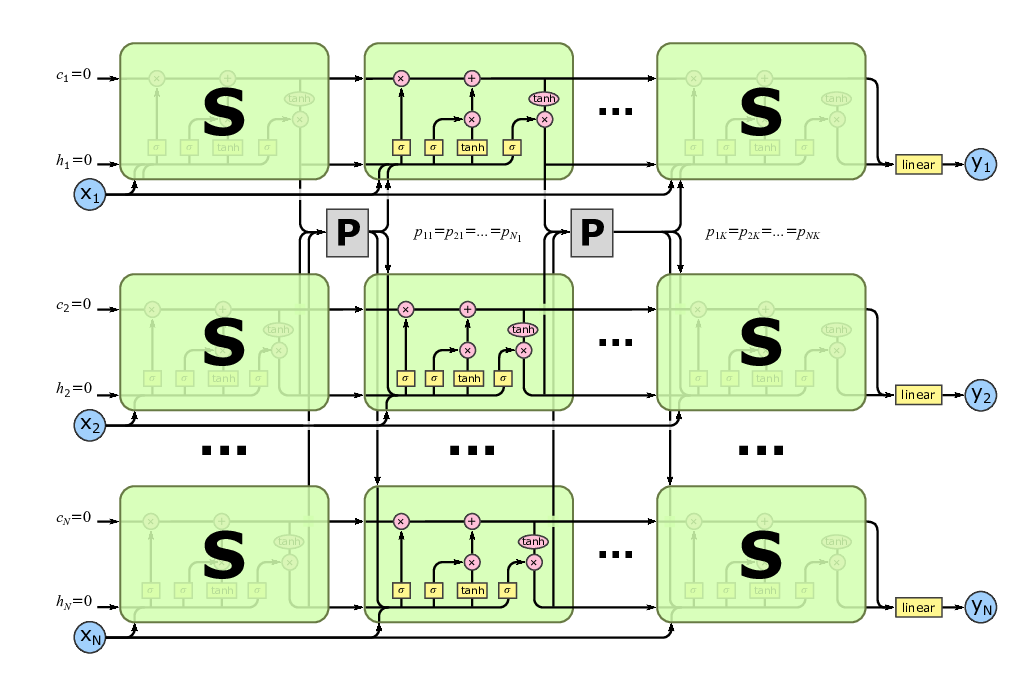} 

\caption{\label{fig:SWARM-cell}\textbf{a)} A SWARM cell processes all entities
individually, much like a LSTM cell, except that additionally the
hidden LSTM states $h_{i,t}$ are pooled appropriately over the all
entities and augment the input of the LSTM cell with population input
$p_{t}$. \textbf{b)} A SWARM layer executes the SWARM cell recurrently
for a certain number of iterations. The LSTM cell and hidden states
are initialized with zero in the first iteration. The input $\mathbf{x}$
is provided in every iteration. After the last iteration, cell and
hidden states of every entity are concateneted along the hidden dimension
and mapped entity-wise with a shared linear layer to the output entities
$\mathbf{y}$. All entities and iterations share the same weights
of the SWARM cell.}
\end{figure}

Our goal was to improve on the limited practical approximation capabilities
of set-linear layers. When we look at their working principle (\ref{eq:set_linear})
then we see that all entities $x_{i}$ are processed individually
with the same affine transformation $\left(\mathbf{W}^{=}-\mathbf{W}^{\neq}\right)x_{i}+\mathbf{b}$
and all entities receive the same additive population update $\mathbf{W}^{\neq}\sum_{i}x_{i}$
. Our idea was to increase expressiveness by letting every single
entity maintain their own memory about how they develop compared to
the development of the whole population during adjacent transformation
steps. The core idea of SWARM mappings is to implement exactly this
entity individual memory. In processing long sequences, it is well
known that gated network architectures like LSTM \citep{hochreiter1997long,gers1999learning}
and GRU \citep{cho2014learning} can carry on information over long
(temporal) distances. In spite of non-temporal but layered architectures,
Highway Networks \citep{DBLP:journals/corr/SrivastavaGS15} have shown
to have the same positive effect on carrying on information through
many adjacent processing steps.

For SWARM we use a modified LSTM cell that receives as input to its
gating networks the entity input $x_{i}$, the last output $h_{i}$,
and additionally a population input $p_{i}=p\left(x_{1},\ldots,x_{N}\right)$,
where $p$ is a set-invariant population function of all entities.
Thus, for the activation of gate $g_{i}$ of entity $i$ , i.e. input,
output, and forget gates (and similarly for the cell update), we have
an additional population term in the activation equation involving
$p_{i}$,

\begin{align*}
g_{i} & =\sigma\left(W_{gx}x_{i}+W_{gh}h_{i}+W_{gp}p_{i}+b_{g}\right)\ .
\end{align*}
 For set-equivariance all parameters have to be shared among all entities,
thus allowing for variable number of entities and permutation invariance.
The update of the memory cell then works as usual in any LSTM. Figure
\ref{fig:SWARM-cell}a) shows an illustration of the SWARM LSTM cell
with population pooling. For processing in a SWARM layer, the cell
is executed in parallel for all entities and repeatedly over several
iterations. During the iterations the input to the cell, $x_{i}$,
will remain the same, but the cell's memory state is constantly updated
with the feedback provided by the population. In the last iteration,
the memory will be sufficient to produce together with the input the
right output $y_{i}$ for entity $i$. Figure \ref{fig:SWARM-cell}b)
depicts a SWARM layer as a recurrent processing unit. Initial values
for $h_{i}$ and $c_{i}$ are set to zero. Taking the SWARM layer
as a set-equivariant building block, nothing speaks against stacking
several of them together or combining them with other set-equivariant
blocks. In our experiments we used one or two layers with a non-linearity
layer between them.

To update population information iteratively is an approach that has
shown to be useful also in related setups. A recurrent application
of set-linear layers in the context of reinforcement learning was
proposed in \citep{sukhbaatar2016learning}. This could be seen as
a SWARM mapping without explicit entity memory and the gating mechanism
introduced by the LSTM cell. After initial experiments, which we had
done with SWARM and a plain RNN cells instead of the LSTM, had shown
difficult training and instable learning behaviour (in particular
for larger populations), we didn't follow this path further.

Iterative LSTM updates are proposed in \citep{vinyals2015order}.
A population vector is iteratively build up with an LSTM cell. In
every iteration, the model attends to an entity embedding and thus
succesively pulls the information of the whole population into a permutation
invariant output vector (which in turn is used to generate an output
set). It is does not the iteration entity-wise, which, however, turns
out beneficial when the primary interest is on permutation-equivariant
mappings.

\section{Experiments\protect\footnote{All data and experiment code can be found at \protect\href{https://github.com/zalandoresearch/SWARM}{https://github.com/zalandoresearch/SWARM} }}

\subsection{\label{subsec:Direct-Amortized-Clustering}Direct Amortized Clustering}

We compare the performance of SWARM layers and other architectures
in an amortized clustering experiment. The model is presented a number
of $N$ entities at a time and its task is to simultaneously assign
every entity to one out of $n_{clust}$ cluster indices. As we are
primarily interested in set-equivariant rather than set-invariant
mappings, we try to learn assignments for data points to clusters
directly. It turned out that this is a rather challenging task that
is difficult for many models to solve. For comparison in, Sec. \ref{subsec:Parameterized-Amortized-Clustering},
we also replicate the exact amortized clustering experiment that was
presented in \citep{DBLP:journals/corr/abs-1810-00825}, where the
parameters of a Mixture of Gaussians have to be estimated instead
the cluster assignments.

Training amortized clustering is a supervised learning task similar
to classification. We want to assign every entity to a class (or cluster).
However, we are confronted with the permutation ambiguity of clustering.
Therefore, we leverage the \emph{Hungarian Algorithm }\citep{kuhn1955hungarian}
to find the optimal assignment of clusters to labels. In theory, there
are $n_{clust}!$ possible assignment permutations. However, the Hungarian
Algorithm is able to find the optimal assignment in polynomial time
which makes this clustering experiment feasible.

\subsubsection{Dataset}

The dataset comprises 10.000 tasks each having random number of $N$
entities, uniformly between $100$ and $1.000$. Entities are points
in $\mathbb{R}^{2}$ that are iid. drawn from a Mixture of Gaussians.
For every task, the number of Gaussian clusters is drawn uniformly
random between 3 and 10, the cluster centers were drawn iid. from
the standard normal distribution. The cluster covariance were drawn
iid. from the inverse Wishart distribution with 4 degrees of freedom
and a scale factor of $0.05\cdot\mathbf{I}$. Cluster assignments
were uniform over the number of clusters. The dataset was sampled
once with a fixed random seed and then used for all experiments. It
was split into 9.000 tasks for training and 1.000 tasks for validation/testing.

\subsubsection{Models}

\begin{figure}
\centering{}\includegraphics[bb=30bp 70bp 785bp 770bp,clip,width=1\textwidth]{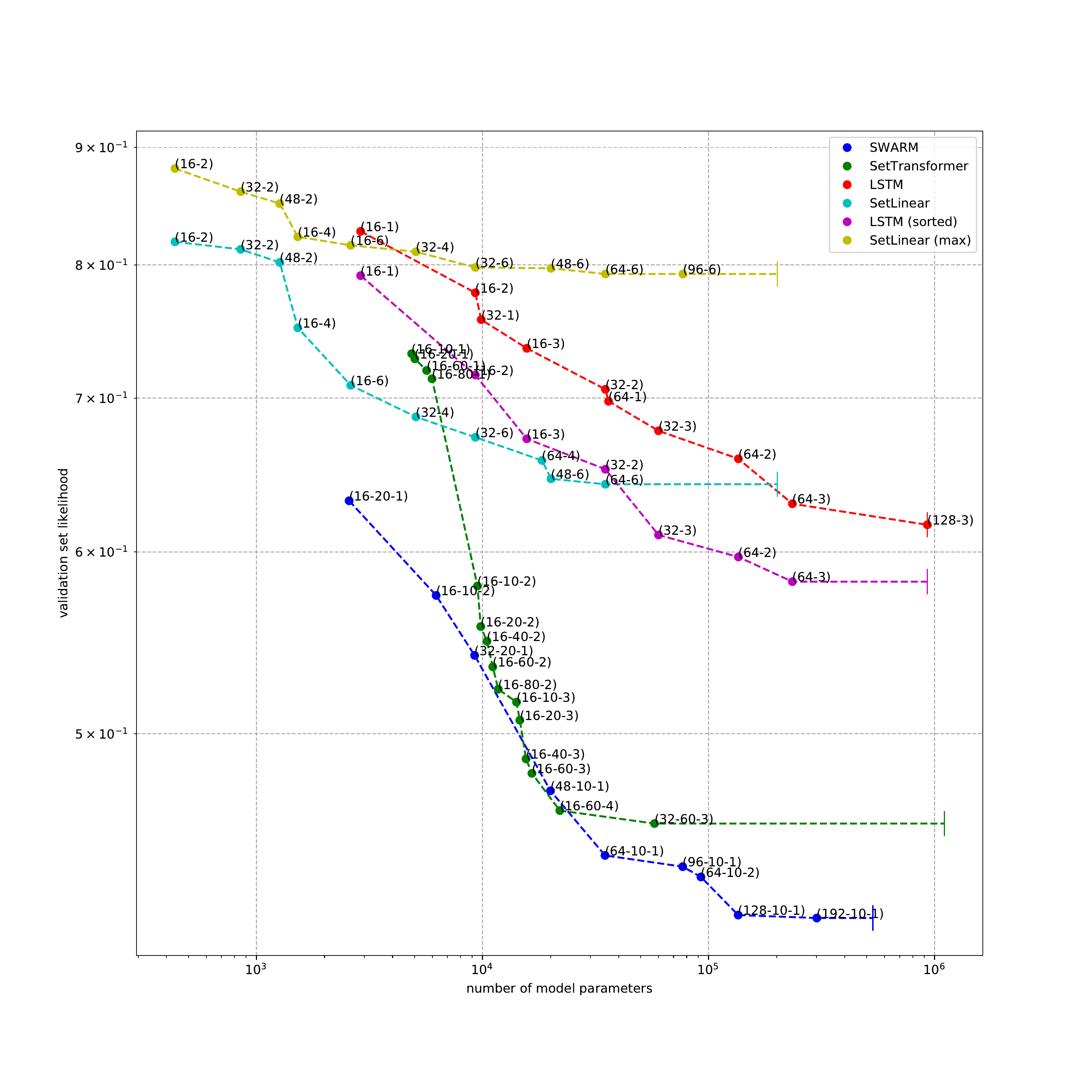}\caption{Validation set performance of different architectures in the direct
amortized clustering experiment. \label{fig:Validation-set-performance}Shown
are only those models that constitute the frontier over performance
and model size. To account for small models that give rise for large
computation overhead, all models were given a fixed computation time
budget.}
\end{figure}

\begin{figure}
\begin{centering}
\includegraphics[bb=16bp 0bp 400bp 260bp,clip,width=0.5\columnwidth]{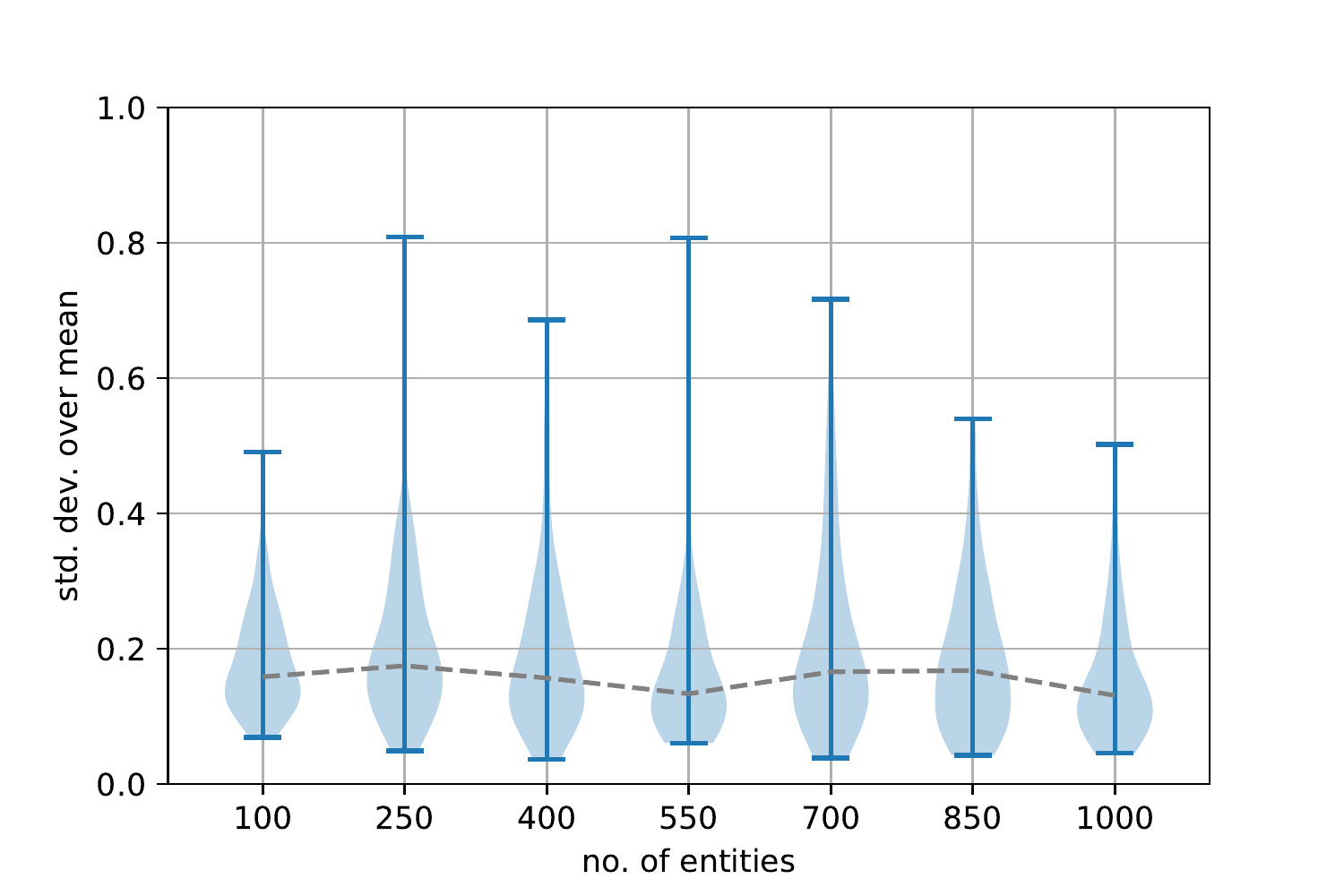}\caption{\label{fig:Randomness-of-predictions}Randomness of predictions made
by the non set-equivariant LSTM model when randomly shuffling the
entities. On average, standard deviation in the clustering loss introduced
by different orderings is 15\% of the expected loss.}
\par\end{centering}
\end{figure}

We compare our approach to different architectures:

\paragraph*{SetLinear and SetLinear (max)}

A feed forward architecture with set-linear mappings (\ref{eq:set_linear})
followed by ReLU non-linearities. We use both (\ref{eq:set_linear})
which performs mean pooling as well a variant with max pooling over
the population.

\paragraph*{LSTM and LSTM (sorted)}

A recurrent architecture based on (potentially multi-layered) bidirectional
LSTM \citep{graves2005framewise} which treats the entities as an
ordered sequence. This is of course no set-equivariant operation.
We have included this model more as a baseline. Also we will use it
to study the effect of an implicit ordering of an otherwise unordered
set. For training of this model we had the entities explicitly shuffled
randomly to avoid that the model learns from any spurious ordering
of the entities. For comparison we also include a variant that does
explicit sorting over the first an then the second dimension of the
entities. Doing so, the LSTM model becomes truly set-equivariant.

\paragraph*{SetTransformer}

These were recently proposed in \citep{DBLP:journals/corr/abs-1810-00825}
and build upon multi-head attention and self attention layers \citep{DBLP:conf/nips/VaswaniSPUJGKP17}.
To be compatible with our setup we only use the encoder part of their
model, which is a set-equivariant function before it gets pooled down
to an output of fixed size in the decoder. From the different architectural
building blocks described in the paper we have chosen the \emph{Induced
Set Attention Blocks (ISAB) }to compare with, as they were reported
best performing. 

\paragraph*{SWARM}

We investigate the proposed SWARM layers as single processing blocks
or stacked on top of each other to form a set-equivariant feed forward
network. When more than one layer was used, a ReLU non-linearity was
used between them.

\subsubsection{Setup}

For all model families, we explored different architecture hyper parameters
to find the best performing model. These were: the number of hidden
units or memory cells, the number of layers, the number of inducing
points in the ISAB blocks, and the number of iterations in the swarm
layer. Table \ref{tab:Model-hyper-parameters} lists the range of
explored hyper parameters and which were applicable for which model. 

Although SWARM architectures give rise to comparably lean architectures,
often with just a single layer, their memory and compute resources
can be demanding depending on the number of iterations performed.
To account for this in performance comparisons, we decided execute
the experiments on fixed compute resources. All models were dedicated
60 minutes net training time\footnote{not including setup, intermediate validation, logging, and checkpoint
generation } on one P100 GPU with two 3.2GHz Xeon CPU cores and 8GB of RAM. Training
was made with batch size 50, no dropout and Adam optimizer.

To automatize and stabilize bulk learning, we used a check pointing
and back tracking heuristic that prevents the model from divergence
due to outliers or too large large learning rates, for example. Whenever
the number of the most recent epochs that all have a better validation
loss than the current one is larger than $\beta=20\%$ of all epochs,
the model weights and the optimizer's internal state are set back
to the checkpoint with the best validation loss so far and the learning
rate is lowered by a factor of $\alpha=0.9$. In orher words, if after
an epoch the optimization would 'thrown back' to a level earlier than
$1-\beta$ from now, the backtracking happens. The first 5 training
epochs, however, are always performed without backtracking.

\begin{table}
\begin{tabular}{l>{\centering}m{1cm}>{\raggedright}p{2cm}>{\raggedright}p{2cm}>{\raggedright}p{2cm}>{\raggedright}p{2cm}}
\hline 
 & 

\textbf{code} & \textbf{a) hidden units} & \textbf{b) inducing points} & \textbf{c) iterations} & \textbf{d) layers}\tabularnewline
\hline 
\hline 
\textbf{SetLinear} & \multirow{2}{1cm}{(a-d)} & \multirow{2}{2cm}{32, 64} & \multirow{2}{2cm}{-} & \multirow{2}{2cm}{-} & \multirow{2}{2cm}{2, 4, 8}\tabularnewline
\textbf{SetLinear (max)} &  &  &  &  & \tabularnewline
\hline 
\textbf{LSTM} & \multirow{2}{1cm}{(a-d)} & \multirow{2}{2cm}{16, 32, 64, 128} & \multirow{2}{2cm}{-} & \multirow{2}{2cm}{-} & \multirow{2}{2cm}{1, 2, 3}\tabularnewline
\textbf{LSTM (sorted)} &  &  &  &  & \tabularnewline
\hline 
\textbf{SetTransformer} & (a-b-d) & 16, 32, 48, 64, 96, 128 & 10, 20, 40, 60, 80 & - & 1, 2, 3, 4\tabularnewline
\hline 
\textbf{SWARM} & (a-c-d) & 16, 32, 48, 64, 96, 128, 192 & - & 2, 5, 10, 20 & 1, 2\tabularnewline
\hline 
\end{tabular} \medskip{}

\caption{\label{tab:Model-hyper-parameters}Overview of the model hyper parameters
that were explored for the direct amortized clustering experiment.
The field ``code'' defines the model codes that are used in Figure
\ref{fig:Validation-set-performance} .}
\end{table}
\begin{table}
\begin{centering}
\begin{tabular}{lcc}
\hline 
 & \textbf{code} & \textbf{validation loss}\tabularnewline
\hline 
\hline 
\textbf{SetLinear} & (64-6) & 0,642 \textpm 0,002\tabularnewline
\textbf{SetLinear (max)} & (96-6) & 0,793 \textpm 0,007\tabularnewline
\hline 
\textbf{LSTM} & (128-3) & 0,617 \textpm 0,008\tabularnewline
\textbf{LSTM (sorted)} & (64-3) & 0,582 \textpm 0,006\tabularnewline
\hline 
\textbf{SetTransformer} & (32-60-3) & 0,457 \textpm 0,015\tabularnewline
\hline 
\textbf{SWARM} & (192-10-1) & \textbf{0,416 \textpm 0,004}\tabularnewline
\hline 
\end{tabular} \medskip{}
\par\end{centering}
\caption{\label{tab:Model-rsults} The best performing architectures the in
the direct amortized clustering experiment for every model class together
with their nos. of parameters and validation loss averaged over indepentent
3 training runs.}
\end{table}

\subsubsection{Results}

Figure \ref{fig:Validation-set-performance} gives an overview of
the performance of the different architectures. It is plotted there
the number of parameters as a measure of model complexity versus the
negative log-likelihood of the model on the validation set. Only models
that are at the performance frontier of model complexity and validation
loss are shown. That means that every architecture that had been explored
but is not shown here would be dominated by another architecture of
the same family with fewer parameters and better validation loss.
Dashed lines in the respective color indicate the empirical frontier
for the different model families. The bar marker at the right of every
frontier indicates the size of the largest model that had been explored.
Next to the models, an architecture code is plotted. See table \ref{tab:Model-hyper-parameters}
for a definition of the architecture codes and the range of hyper
parameter explorations. Table \ref{tab:Model-rsults} shows the best
performing models of each family.

One can clearly see that SWARM layer models outperform all other model
classes and dominate large parts of the overall frontier. Interestingly,
top performing SWARM models have just one layer. This approves empirical
findings of ours from before this study. In an early approach, we
had tried larger stacks of layers or even different SWARM cells in
every iteration. However, none of these higher parametrized architectures
worked particularly well. 

The Set Transformer models dominate the overall frontier slightly
with models between $\approx$15k and 20k parameters. However, for
larger architectures they fail to reach the performance of 1-layer
SWARM models and also for very small models, SWARM is clearly dominating
the frontier.

Clearly worse performing are LSTM and set-linear models, where the
max-pooling in a set-linear model appears to be harmful. Interestingly,
an LSTM based model with explicit sorting performs significantly better
than without, even though it does not reach SWARM or Set Transformer\textquoteright s
performance. 

\begin{figure}
\includegraphics[bb=25bp 30bp 390bp 385bp,clip,width=0.5\columnwidth]{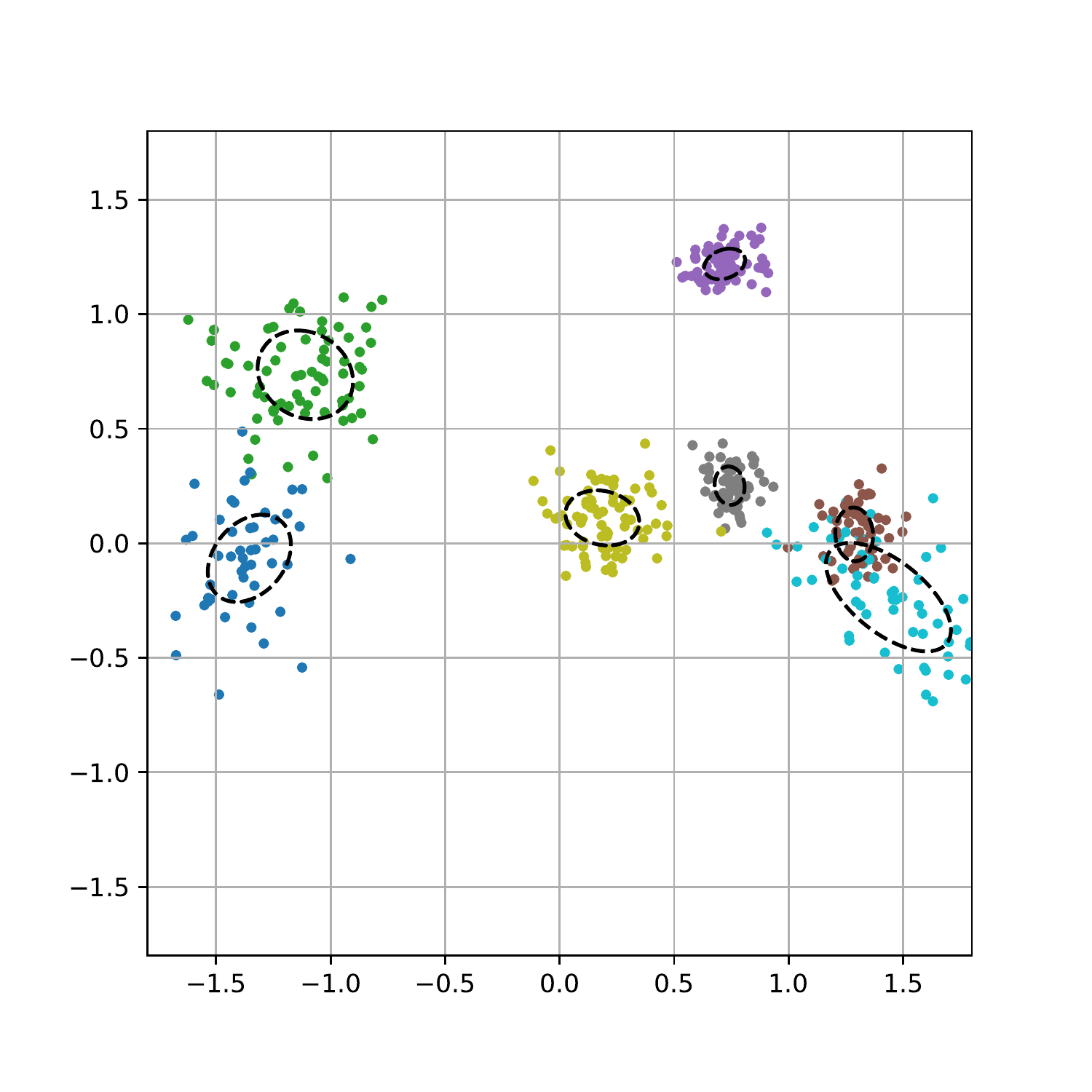}\includegraphics[bb=25bp 30bp 390bp 385bp,clip,width=0.5\columnwidth]{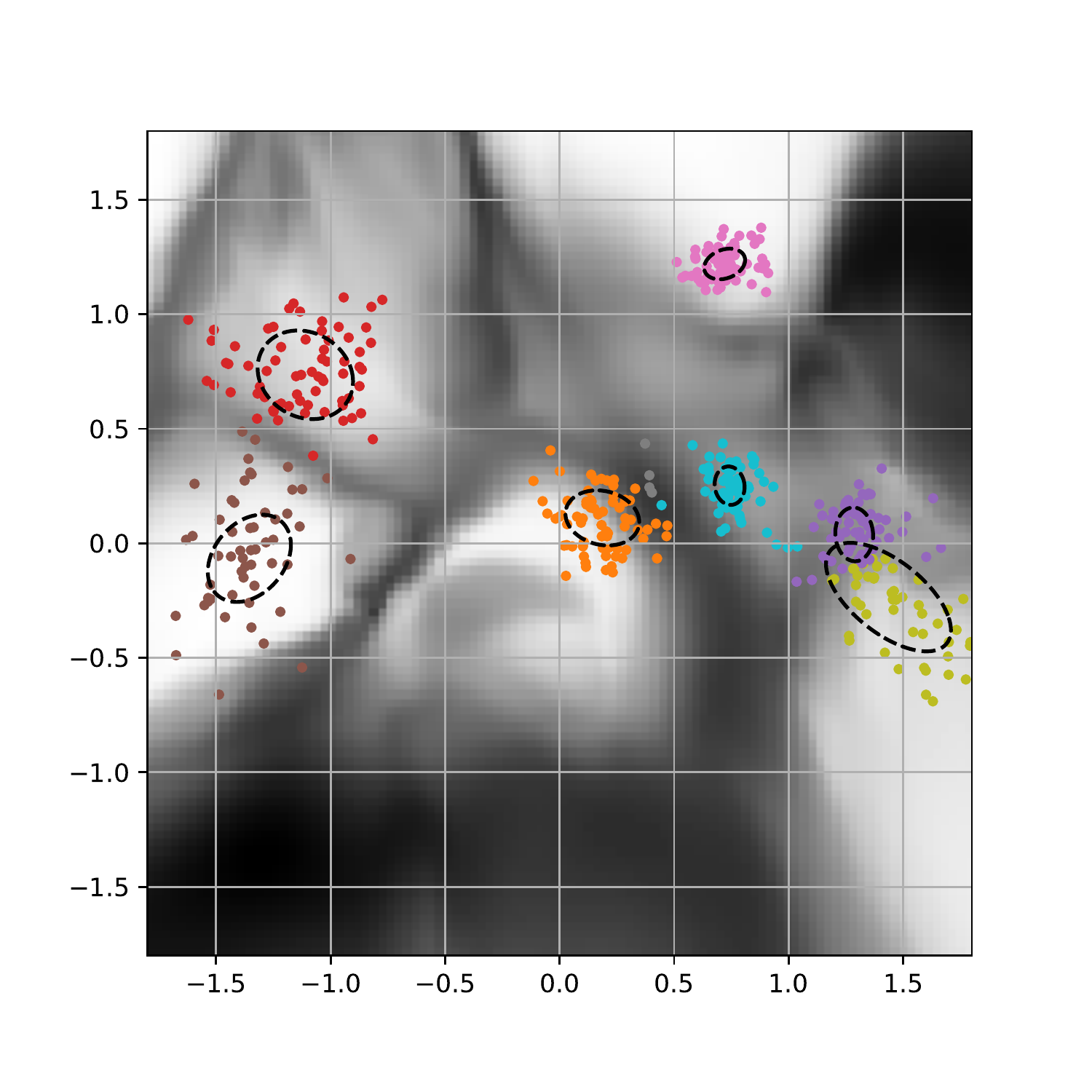}\caption{\label{fig:Example-clustering-task.}Example clustering task. \emph{Left:}
random example task with 500 entities and ground truth cluster assignment
(color coded). \emph{Right:} amortized cluster assignment made by
a model with 1 SWARM layer, 192 memory cells doing 10 iterations.
The gray shades indicate the confidence of cluster assignment at any
position for a hypothetical additional entity joining the population.
It is given by the entropy of the cluster assignment probability (see
Fig. \ref{fig:Cluster-assignment-probability}).}
\end{figure}

\begin{figure}
\includegraphics[bb=130bp 50bp 1050bp 445bp,clip,width=1\columnwidth]{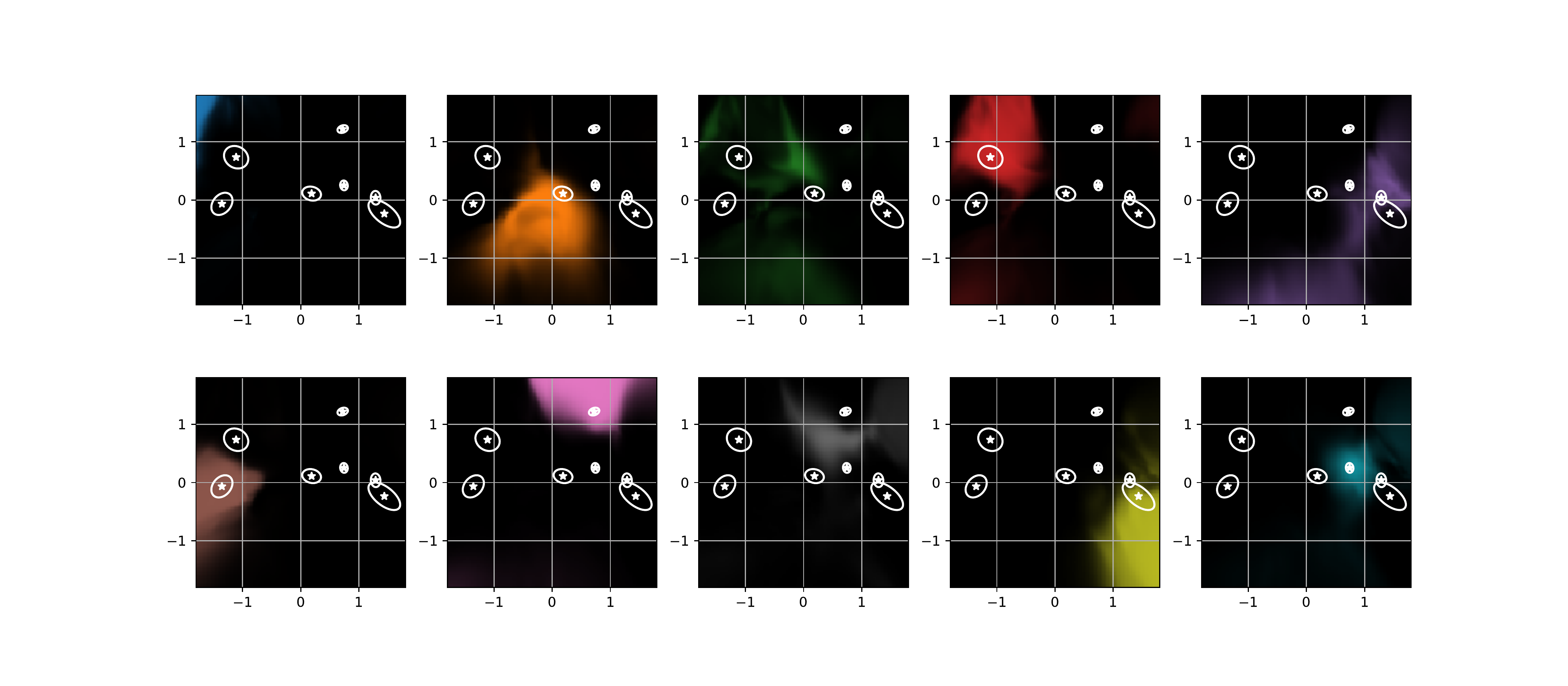}\caption{\label{fig:Cluster-assignment-probability}Cluster assignment probability
(black: $p=0$, full color: $p=1$) of a new hypothetical data point
joining the population. Each panel shows one of 10 possible clusters.
The white ellipses indicate the ground truth Gaussian distribution.
The spare blue, green, and gray clusters only cover regions with no
data points, as this task had only 7 clusters.}
\end{figure}

Results of the amortized clustering generated by a single SWARM layer
with 192 units and 10 iteration can be seen in Figure \ref{fig:Example-clustering-task.}.
The left panel shows the ground truth data generated from a test set.
This task has 7 clusters shown in different colors together with the
covariance ellipses of the generating Gaussian distributions. The
right panel shows the clusters assigned by the SWARM layer. Note that
the colors don't map one by one because of the above mentioned permutation
ambiguity in clustering. Apart from the region on the right with overlapping
clusters the model's cluster assignment is quite consistent. The gray
shaded area shows the assignment confidence for a single entity $x^{*}$
that was augmented to the population $\mathbf{x}$ resulting in $\hat{\mathbf{x}}=\left\{ x_{i}\right\} _{i=1\ldots N}\cup\left\{ x^{*}\right\} $.
Being $\hat{\mathbf{y}}$ the logits after the transformation with
the SWARM layer, $\hat{\mathbf{y}}=f_{SWARM(192-10-1)}\left(\mathbf{\hat{x}}\right)$,
the gray level corresponds to the entropy at the respective position
$x^{*}$. Darker regions are regions of higher entropy, thus lower
assignment confidence. Note that in regions where the clustering is
more ambiguous the entropy is also comparably large. Interestingly,
also unpopulated regions in the input space can lead to regions of
high confidence (for example, the pink cluster in Fig. \ref{fig:Example-clustering-task.},
right). Similar figures emerge from the other models too. Figure \ref{fig:Cluster-assignment-probability}
shows the cluster assignment probabilities for a new point $x^{*}$.
Also the 3 unused clusters (blue, green, and gray) dominate some unpopulated
regions in input space where they would \textquotedbl{}pick up\textquotedbl{}
new data points in new clusters.

Finally, we had a closer look at the performance of the LSTM models.
Figure \ref{fig:Validation-set-performance} shows that explicit sorting,
i.e. making the problem permutation equivariant explicitly, helps
to improve their average performance to some degree. However, the
variance of clustering results can be quite large due to different
orderings of the population. To show this, for populations from 100
to 1.000 entities, clustering losses of the best unsorted LSTM model
(128-3) were recorded for 100 clustering tasks and 1.000 different
random shufflings of the entities per task. Figure \ref{fig:Randomness-of-predictions}
shows a violin distribution plot of the standard deviations per task,
scaled by the average loss of that task. We see that it is in the
range of 15\% (but can be up to 80\% for certain tasks), which is
quite significant. Note that for the set-equivariant models this value
is zero by construction.

\subsection{\label{subsec:Parameterized-Amortized-Clustering}Parametrized Amortized
Clustering}

In the direct amortized clustering experiment of section \ref{subsec:Direct-Amortized-Clustering},
it showed that Set Transformer architectures are the strongest competitor
for SWARM mappings. Initially they were introduced in \citep{DBLP:journals/corr/abs-1810-00825}
in a permutation-invariant setting, where a permutation equivariant
encoder is coupled with a pooling or decoder block. Among others,
they also present a parametrized amortized clustering experiment,
where instead of direct cluster assignments the parameters of a mixture
of Gaussians have to be estimated. For comparison we reproduce this
experiment here and show that SWARM mappings with simple average pooling
are on par with Set Transformers in this task.

For clustering the model must generate for every task a permutation-invariant
parameter vector of dimension $d=n_{clust}\left(2n_{dim}+1\right)$.
For our experiments we used two approaches: (i) a SWARM layer with
output size $d$ is average pooled over all entities, and (ii) a SWARM
layer of output size 128 feeds into a\emph{ Pooling by Multihead Attention
(PMA)} block as defined in \citep{DBLP:journals/corr/abs-1810-00825}.

The dataset is simpler than our's in section \ref{subsec:Direct-Amortized-Clustering}.
It comprises a fixed number of 4 isotropic Gaussian clusters in two
dimensions with constant standard deviation $\sigma=0.3$. The cluster
means are distributed uniformly in the interval $(-4,4)$ in both
dimensions, and the number of data points for every task is drawn
uniformly from the interval $[100,500]$. The cluster weights are
different in every task and are drawn from a flat Dirichlet distribution.
To be comparable, we used the same training procedure as in \citep{DBLP:journals/corr/abs-1810-00825}
with a total of 500.000 random tasks, none of them used twice. At
70\% of the iterations, as well we reduce the learning rate be a factor
of $0.1$. Since it is not clear which loss the authors report, we
show both the average training log-likelihood of the last epoch and
a test log-likelihood of another 10.000 tasks that were not used during
training. Results are summarized in table \ref{tab:Prameterized-amortized-clusterin}.
It shows that a single SWARM layer (768-20-1) with pooling in on par
with the best Set Transformer model. Interestingly, while for the
Set Transformers the PMA decoder is crucial for performance, in SWARM
models it seems to be obstructive. Results are constantly worse when
using PMA instead of simple average pooling.

\begin{table}
\centering{}%
\begin{tabular}{cccccc}
\hline 
 & \textbf{Architecture} & \multicolumn{2}{c}{\textbf{pooling}} & \multicolumn{2}{c}{\textbf{PMA}}\tabularnewline
\hline 
\hline 
\multirow{4}{*}{\begin{turn}{90}
\textbf{Set Tr.}
\end{turn}} & SAB & \multicolumn{2}{c}{-1.6772 \textpm{} 0.0066} & \multicolumn{2}{c}{-1.5145 \textpm{} 0.0046}\tabularnewline
 & ISAB (16) & \multicolumn{2}{c}{-1.6955 \textpm{} 0.0730} & \multicolumn{2}{c}{-1.5009 \textpm{} 0.0068}\tabularnewline
 & \textbf{ISAB (32)} & \multicolumn{2}{c}{-1.6353 \textpm{} 0.0182} & \multicolumn{2}{c}{\textbf{-1.4963 \textpm{} 0.0064}}\tabularnewline
 & ISAB (64) & \multicolumn{2}{c}{-1.6349 \textpm{} 0.0429} & \multicolumn{2}{c}{-1.5042 \textpm{} 0.0158}\tabularnewline
\hline 
\multirow{7}{*}{\begin{turn}{90}
\textbf{SWARM}
\end{turn}} &  & test  & train & test & train\tabularnewline
 & (256-10-1) & -1.5205\textpm 0.0040 & -1.5179\textpm 0.0053 & -1.5237\textpm 0.0046 & -1.5221\textpm 0.0062\tabularnewline
 & (256-20-1) & -1.5235\textpm 0.0032 & -1.5226\textpm 0.0053 & -1.5222\textpm 0.0098 & -1.5236\textpm 0.0079\tabularnewline
 & (512-10-1) & -1.5109\textpm 0.0064 & -1.5112\textpm 0.0061 & -1.5131\textpm 0.0034 & -1.5127\textpm 0.0021\tabularnewline
 & (512-20-1) & -1.5065\textpm 0.0067 & -1.5096\textpm 0.0093 & -1.5172\textpm 0.0070 & -1.5154\textpm 0.0096\tabularnewline
 & (768-10-1) & -1.5034\textpm 0.0064 & -1.5031\textpm 0.0101 & -1.5194\textpm 0.0116 & -1.5158\textpm 0.0077\tabularnewline
 & \textbf{(768-20-1)} & \textbf{-1.4986\textpm 0.0053} & \textbf{-1.4974\textpm 0.0035} & -1.5141\textpm 0.0078 & -1.5128\textpm 0.0097\tabularnewline
\hline 
\end{tabular}\medskip{}
\caption{\label{tab:Prameterized-amortized-clusterin}Parametrized amortized
clustering results for different encoder architectures and pooling
and PMA decoders. Set Transformer results are taken from \citep{DBLP:journals/corr/abs-1810-00825}. }
\end{table}

\subsection{SWARM Transformer}

\begin{figure}
\includegraphics[bb=40bp 60bp 325bp 228bp,clip,width=0.5\columnwidth]{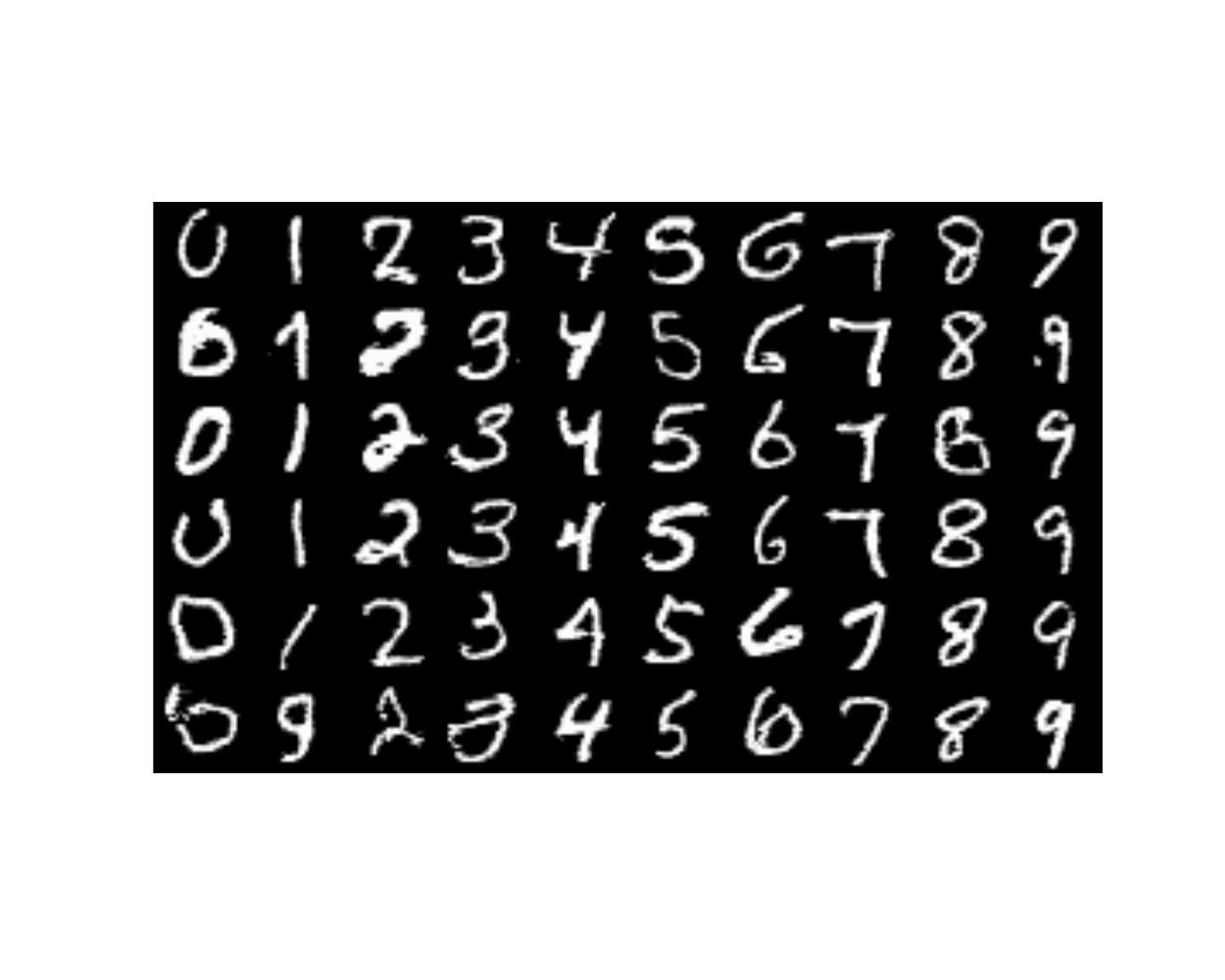}\includegraphics[bb=40bp 60bp 325bp 228bp,clip,width=0.5\columnwidth]{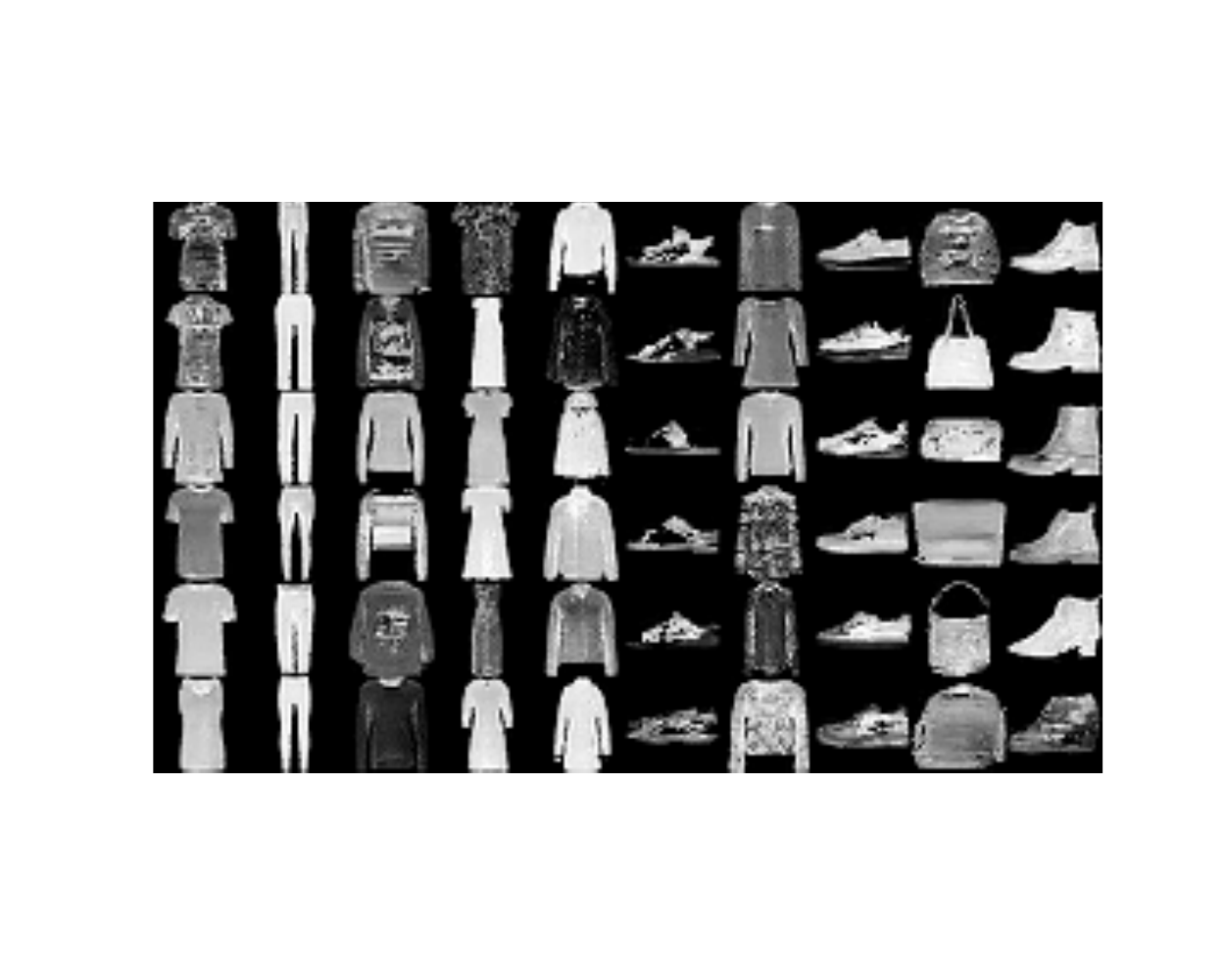}\\
\includegraphics[bb=40bp 60bp 325bp 228bp,clip,width=0.5\columnwidth]{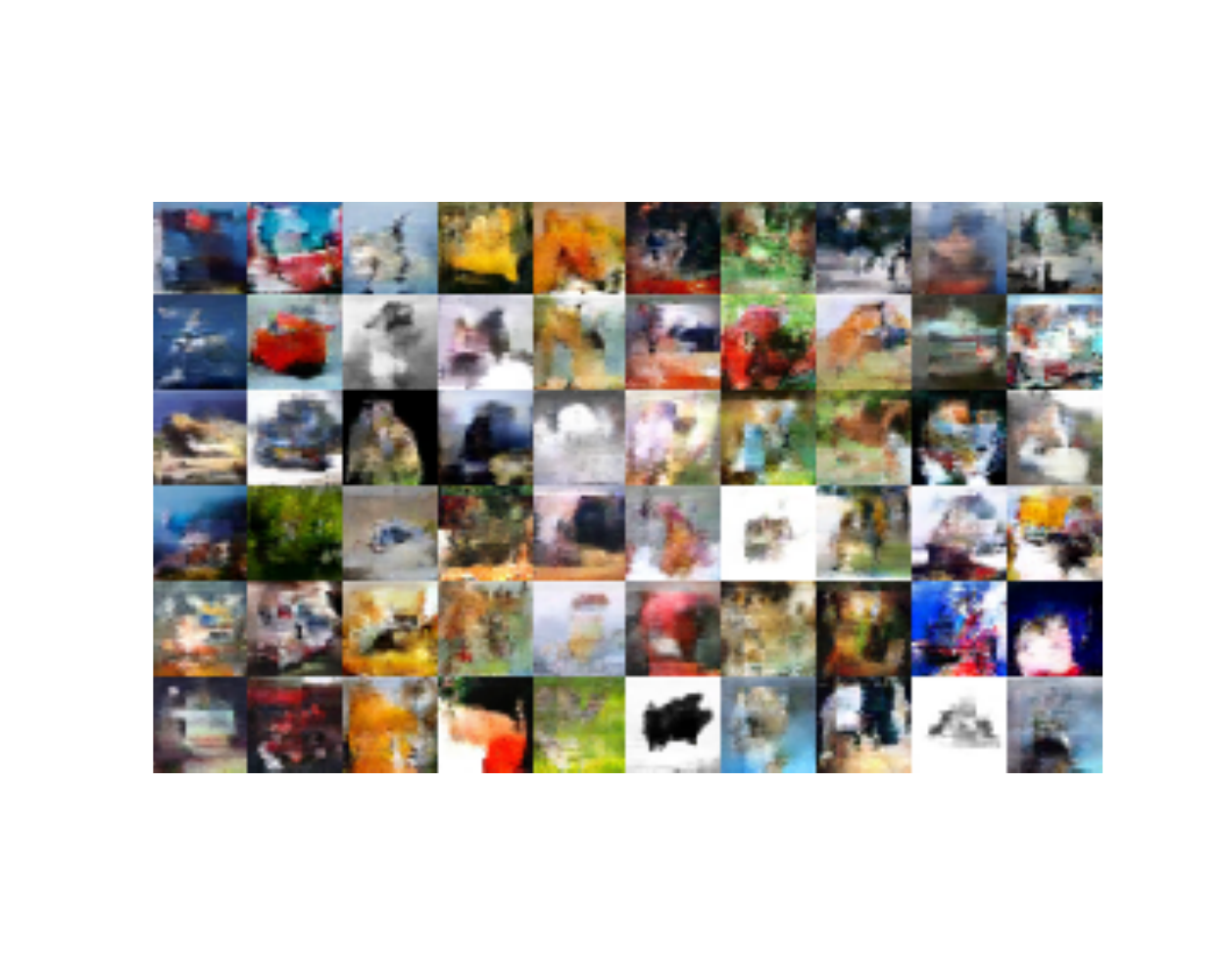}\hfill{}%
\begin{tabular}[b]{cccc}
\hline 
dataset & \multicolumn{2}{c}{parameters} & b/dim\tabularnewline
\hline 
\hline 
{\small{}MNIST} & (256-10-1) & 980k & 0,89\tabularnewline
{\small{}FashionMNIST} & (256-10-1) & 980k & 2,84\tabularnewline
{\small{}CIFAR10} & (512-20-1) & 3,5M & 3,93\tabularnewline
\hline 
 &  &  & \tabularnewline
 &  &  & \tabularnewline
\end{tabular}\caption{Class conditional samples of images generated with a 1-layer SWARM
Transformer. }
\end{figure}

This section presents preliminary results of SWARM layers in an interesting
and slightly different setting. A set-equivariant layer is the main
building block for powerful neural network architectures, which recently
enjoy increasing popularity - Transformers \citep{DBLP:conf/nips/VaswaniSPUJGKP17}.
Scaled Dot Product Attention, Self Attention, Multi-Head Self Attentions
are the ingredients for several models that are state of the art in
many challenges currently. Surprisingly, the set-equivariance is not
actually needed there. To be precise, it is even explicitly eliminated
by the introduction of positional encodings. Still, reportedly transformers
frequently outperform recurrent or convolutional architectures. The
question was, could a SWARM layer also be used in a transformer-like
setting. We investigate this with the task of image generation, as
there have been reported great success with transformers recently
\citep{parmar2018image,child2019generating}. We have adopted the
setup widely from the Image Transformer. To build a \emph{SWARM Transformer
}we had to replace the pooling operation in the SWARM layer with a
causal mean pooling, that is $p_{i}=\frac{\sum_{i'<i}h_{i}}{\sum_{i'<i}1}$
where the entities are explicitly ordered along the scan lines of
the image. We further used 256 dimensional fixed positional encodings,
similar to those in the Image Transformer, and 256 dimensional trainable
input and channel embeddings. As they are adjustable and are immediately
followed by a linear layer operation in the SWARM cell, instead of
concatenating them, we have added them up as suggested in \citep{DBLP:conf/nips/VaswaniSPUJGKP17}.
SWARM Transformer generated MNIST \citep{lecun1998gradient} and FashionMNIST
\citep{xiao2017fashion} samples look very convincing and also their
likelihoods are state of the art (cf. \citep{nalisnick2018deep}).
The CIFAR10 \citep{krizhevsky2009learning} results are more off (cf.
the survey in \citep{DBLP:journals/corr/OordKVEGK16} and \citep{parmar2018image})
and also the samples are less visually appealing. We hope that with
refined architectures can improve on that.

\section{Conclusion}

We have presented a powerful yet simple architecture for set-equivariant
functions and could demonstrate that it outperforms other state of
the art models in an amortized clustering experiment. There, hyper
parameter exploration gave rise to best performing SWARM architectures
that are particularly simple. Best models have just one layer and
don't require dropout, layer normalization, or other training stabilizing
measures.

Notably, the SWARM layers can be used as an immediate replacement
for attention and self-attention blocks if the pooling function is
designed appropriately. We could demonstrate that this can yield state
of the art performance in image-transformer-like tasks (MNIST and
FashionMNIST) with much simpler architectures than attention based
image transformers. For our future work it remains to systematically
analyze in which areas SWARM mappings are beneficial over attention
based models. In particular we want to better understand why the SWARM
transformer performed in our experiment so much better on the 1-channel
tasks MNIST and Fashion MNIST compared to CIFAR10. 

\bibliographystyle{plain}
\bibliography{citations,IEEEfull}

\end{document}